\documentclass[10pt,twocolumn,letterpaper]{article}

\usepackage{cvpr}              

\usepackage{graphicx}
\usepackage{amsmath}
\usepackage{amssymb}
\usepackage{booktabs}
\usepackage[table,xcdraw]{xcolor}

\usepackage{multirow}
\usepackage[normalem]{ulem}
\useunder{\uline}{\ul}{}
\DeclareMathOperator*{\argmax}{argmax}

\usepackage[pagebackref,breaklinks,colorlinks]{hyperref}

\usepackage[capitalize]{cleveref}
\crefname{section}{Sec.}{Secs.}
\crefname{section}{Section}{Sections}
\crefname{table}{Table}{Tables}
\crefname{table}{Tab.}{Tabs.}

\def\cvprPaperID{473} 
\def\confName{CVPR}
\def\confYear{2023}

\begin{document}

\title{Long-Tailed Visual Recognition via Self-Heterogeneous Integration with Knowledge Excavation}



\author{Yan Jin$^{1,2}$ \qquad Mengke Li$^3$ \qquad Yang Lu$^{1,2}$\thanks{Corresponding author: Yang Lu, luyang@xmu.edu.cn} \qquad Yiu-ming Cheung$^4$ \qquad Hanzi Wang$^{1,2}$ \\
$^1$ \small{Fujian Key Laboratory of Sensing and Computing for Smart City, School of Informatics, Xiamen University, Xiamen, China}\\
$^2$ \small{Key Laboratory of Multimedia Trusted Perception and Efficient Computing,}\\ 
\small{Ministry of Education of China, Xiamen University, Xiamen, China}\\
$^3$ \small{Guangdong Laboratory of Artificial Intelligence and Digital Economy (SZ), Shenzhen, China}\\
$^4$ \small{Department of Computer Science, Hong Kong Baptist University, Hong Kong SAR, China}\\
\hspace{-7px}{\tt\small jinyan7973@gmail.com limengke@gml.ac.cn \{luyang, Hanzi.Wang\}@xmu.edu.cn ymc@comp.hkbu.edu.hk}
}

\maketitle

\begin{abstract}
Deep neural networks have made huge progress in the last few decades. 
However, as the real-world data often exhibits a long-tailed distribution, vanilla deep models tend to be heavily biased toward the majority classes. 
To address this problem, state-of-the-art methods usually adopt a mixture of experts (MoE) to focus on different parts of the long-tailed distribution.
Experts in these methods are with the same model depth, which neglects the fact that different classes may have different preferences to be fit by models with different depths.
To this end, we propose a novel MoE-based method called Self-Heterogeneous Integration with Knowledge Excavation (SHIKE).
We first propose Depth-wise Knowledge Fusion (DKF) to fuse features between different shallow parts and the deep part in one network for each expert, which makes experts more diverse in terms of representation. 
Based on DKF, we further propose Dynamic Knowledge Transfer (DKT) to reduce the influence of the hardest negative class that has a non-negligible impact on the tail classes in our MoE framework. 
As a result, the classification accuracy of long-tailed data can be significantly improved, especially for the tail classes.
SHIKE achieves the state-of-the-art performance of 56.3\%, 60.3\%, 75.4\% and 41.9\% on CIFAR100-LT (IF100), ImageNet-LT, iNaturalist 2018, and Places-LT, respectively. The source code is available at
\href{https://github.com/jinyan-06/SHIKE}{{https://github.com/jinyan-06/SHIKE}}.

\end{abstract}

\section{Introduction}
\label{sec:intro}      
Deep learning has made incredible progress in visual recognition tasks during the past few years. 
With well-designed models, e.g., ResNet~\cite{he2016deep}, and Transformer~\cite{vaswani2017attention}, deep learning techniques have outperformed humans in many visual applications, like image classification~\cite{krizhevsky2017imagenet}, semantic segmentation~\cite{girshick2014rich, long2015fully}, and object detection~\cite{ren2015faster, szegedy2013deep}. 
One key factor in the success of deep learning is the availability of large-scale datasets~\cite{deng2009imagenet, zhou2017places, van2018inaturalist}, which are usually manually constructed and annotated with balanced training samples for each class. 
However, in real-world applications, data typically follows a long-tailed distribution, where a small fraction of classes possess massive samples, but the others are with only a few samples~\cite{cui2019class, kang2020exploring, menon2020long, liu2019large, Cao2019Learning}.
Such imbalanced data distribution leads to a significant accuracy drop for deep learning models trained by empirical risk minimization (ERM)~\cite{vapnik1991principles} as the model tends to be biased towards the head classes and ignore the tail classes to a great extent.
Thus, the model's generalization ability on tail classes is severely degraded.

\begin{figure}[t]
    \centering
    \includegraphics[width=1.0\linewidth]{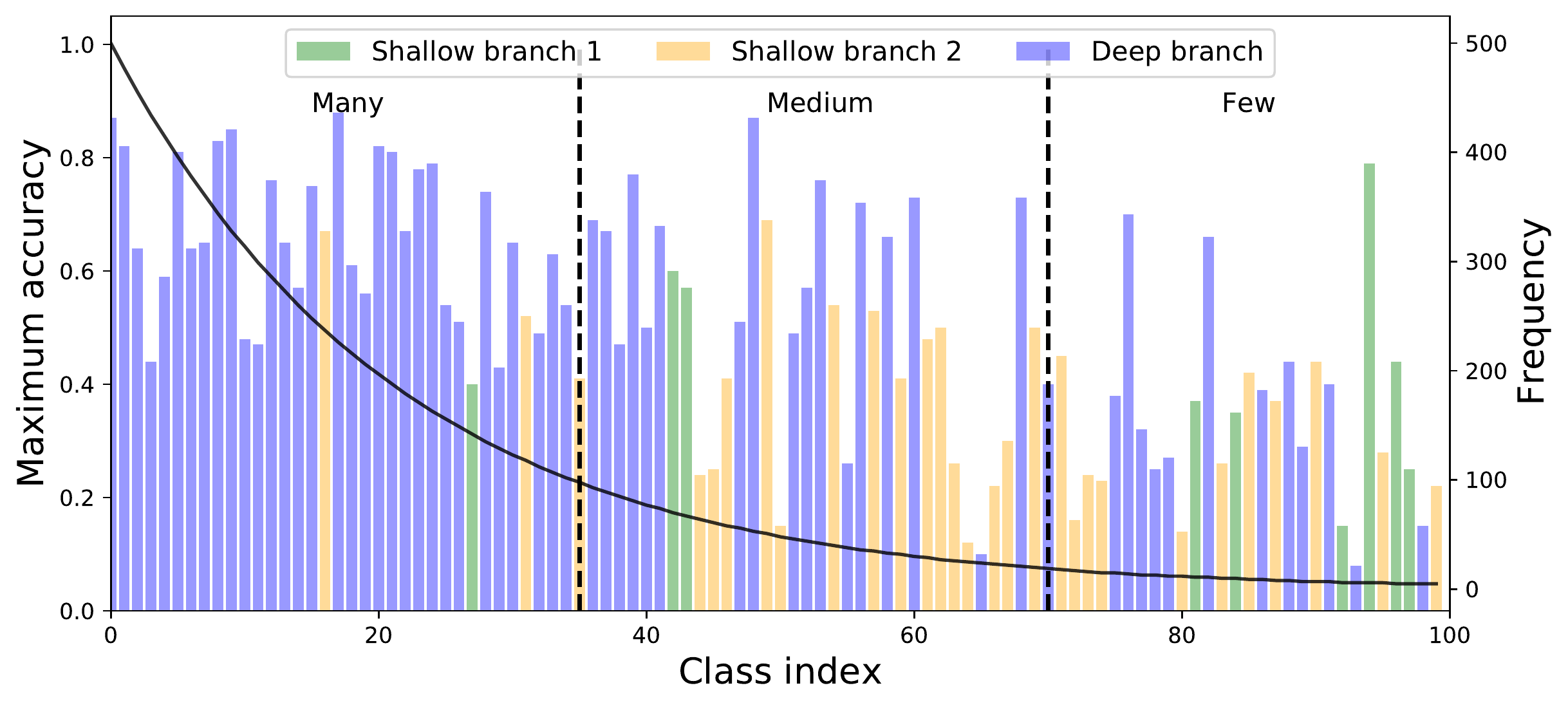}
    \vspace{-16px}
    \caption{Comparison of test accuracy of a ResNet-32 model with two shallow branches and a deep branch. The model is jointly trained on CIFAR100-LT with an imbalance factor of 100. Only the highest accuracy among the three branches is shown for each class.}
    \label{motivation}
\vspace{-10px}
\end{figure}

The most straightforward action for long-tailed recognition often focuses on re-balancing the learning process from either a data processing~\cite{Park2022Majority, Buda2018systematic, kang2019decoupling} or cost-sensitive perspective~\cite{zhou2005training, Khan2018cost, cui2019class}. 
Recently, methods proposed for long-tailed data have drawn more attention to representation learning.
For example, the decoupling strategy~\cite{kang2019decoupling} is proposed to deal with the inferior representation caused by re-balancing methods. 
Contrastive learning~\cite{kang2020exploring, cui2021parametric} specializes in learning better and well-distributed representations. 
Among them, the methods that achieve state-of-the-art performance are usually based on a mixture of experts (MoE), also known as multi-experts.
Some MoE-based methods prompt different experts to learn different parts of the long-tailed distribution (head, medium, tail)~\cite{xiang2020learning, li2020overcoming, cai2021ace, cui2022reslt}, while some others were designed to reduce the overall model's prediction variance or uncertainty~\cite{wang2020long, li2022trustworthy}. 

Unlike traditional ensemble learning methods that adopt independent models for joint prediction, the MoE-based methods for long-tailed learning often adopt a multi-branch model architecture with shared shallow layers and exclusive deep layers.
Thus, the features generated by different experts are actually from the model with the same depth, although the methods force them to be diverse from various perspectives.
Recently, self-distillation \cite{zhang2021self} is proposed to enable shallow networks to have the ability to predict certain samples in the data distribution.
This brings us to a new question: can we integrate the knowledge from shallow networks into some experts in MoE to fit the long-tailed data in a self-adaptive manner regardless of the number of samples?
With this question, we conduct a quick experiment to reveal the preference of the deep neural network on different classes in long-tailed data. 
A ResNet-32 model with branches directly from shared layers is adopted. 
Each branch contains an independent classifier after feature alignment, and all classifiers are re-trained with balanced softmax cross entropy \cite{ren2020balanced}.
\cref{motivation} shows the highest accuracy among the three branches for each class.
We can clearly observe that shallow parts of the deep model are able to perform better on certain tail classes.
This implies that different parts of the long-tailed distribution might accommodate the network differently according to the depth. 
Thus the shallow part of the deep model can provide more useful information for learning the long-tailed distribution. 

Driven by the observation above, we propose a novel MoE-based method called Self-Heterogeneous Integration with Knowledge Excavation (SHIKE). 
SHIKE adopts an MoE-based model consisting of heterogeneous experts along with knowledge fusion and distillation.
To fuse the knowledge diversely, we first introduce Depth-wise Knowledge Fusion (DKF) as a fundamental component to incorporate different intermediate features into deep features for each expert. 
The proposed DKF architecture can not only provide more informative features for experts but also optimize more directly to shallower layers of the networks by mutual distillation. 
In addition, we design Dynamic Knowledge Transfer (DKT) to address the problem of the hardest negatives during knowledge distillation between experts.
DKT elects the non-target logits with large values to reform non-target predictions from all experts to one grand teacher, which can be used in distilling non-target predictions to suppress the hardest negative, especially for the tail classes.
DKT can fully utilize the structure of MoE and diversity provided by DKF for better model optimization.
In this paper, our contributions can be summarized as follow: 
\begin{itemize}
    \item We propose Depth-wise Knowledge Fusion (DKF) to encourage feature diversity in knowledge distillation among experts, which releases the potential of the MoE in long-tailed representation learning. 
    \item We propose a novel knowledge distillation strategy DKT for MoE training to address the hardest negative problem for long-tailed data, which further exploits the diverse features fusing enabled by DKF.
    \item We outperform other state-of-the-art methods on four benchmarks by achieving performance 56.3\%, 60.3\%, 75.4\% and 41.9\% accuracy for CIFAR100-LT (IF100), ImageNet-LT, iNaturalist 2018, and Places-LT, respectively.
\end{itemize}

\section{Related work}
\paragraph{Long-tailed Visual Recognition.} 
Long-tailed visual recognition aims at improving the accuracy of the tail classes with the least influence on the head classes.
Re-sampling is the most common practice in early methods for long-tailed learning, which mainly focuses on balancing the data distribution during model training \cite{chawla2002smote, kang2019decoupling, mahajan2018exploring, wang2020devil, zang2021fasa}. 
In terms of model optimization, re-weighting aims to re-balance classes in the way of adjusting loss value for different classes during training~\cite{elkan2001foundations, hong2021disentangling, lin2017focal, park2021influence, ren2020balanced, zhang2021distribution, zhou2005training}.
Data augmentation enables balanced training by means of either transferring the information from the head classes to the tail classes~\cite{kim2020m2m, wang2021rsg} or generating data for the tail classes using prior~\cite{zang2021fasa} or estimated statistics~\cite{li2021metasaug}. 
 
Some other methods adopt logit adjustment to calibrate data distribution by post-hoc shifting the logit based on label frequencies in order to obtain a large margin between classes~\cite{hong2021disentangling, li2022key, li2022long, menon2020long, provost2000machine, zhang2021distribution}.
As the re-balancing methods usually promote the accuracy of tail classes at the cost of harming the head classes, decoupled training~\cite{kang2019decoupling} and contrastive learning methods~\cite{cui2021parametric, zhu2022balanced} are proposed to learn the better representation for long-tailed learning.


More recently, methods based on the mixture of experts (MoE) have been explored to improve performance by integrating more than one model in the learning framework.
The basic idea is to make the experts focus on different parts of the long-tailed data.
BBN~\cite{zhou2020bbn} is proposed to use a two-branched to learn the long-tailed distribution and the balanced distribution simultaneously along with a smooth transition between them. 
BAGS~\cite{li2020overcoming} and LFME~\cite{xiang2020learning} reduce the related imbalance ratio by dividing the long-tailed distribution into several sub-groups with several experts fitting on them.
ACE~\cite{cai2021ace} and ResLT~\cite{cui2022reslt} allow experts to be skilled at different parts of the long-tailed distribution and to complement each other.
RIDE~\cite{wang2020long} and TLC~\cite{li2022trustworthy} utilize several experts to learn the long-tailed distribution independently. 
Thus, the predictions of all experts are gradually integrated to reduce the overall model variance or uncertainty.
NCL~\cite{li2022nested} adopts several complete networks to learn the long-tailed distribution individually and collaboratively along with self-supervised contrastive strategy~\cite{cui2021parametric}.
\vspace{-10px}
\paragraph{Knowledge Distillation.} Knowledge Distillation (KD) \cite{hinton2015distilling} is originally proposed to transfer knowledge from a large teacher model to a small student model by using soft labels output from the large model as targets. 
There are two main KD directions: logit-based KD and feature-based KD.
Logit-based KD methods~\cite{kim2018paraphrasing, furlanello2018born, zhang2018deep, cho2019efficacy, mirzadeh2020improved} directly use the output of the teacher model as supervision to guide the student model, while feature-based KD methods~\cite{romero2014fitnets, zagoruyko2016paying, huang2017like, kim2018paraphrasing, ahn2019variational, deng2009imagenet, heo2019knowledge,  heo2019comprehensive, tian2019contrastive, tung2019similarity} is designed to match the intermediate features between the teacher model and the student model.
Recently, self-distillation has been proposed to utilize the idea of knowledge distillation for better and more efficient model optimization \cite{zhang2021self}.
It treats the deep model as the teacher model to transfer knowledge directly to shallow layers viewed as student models.

\begin{figure*}[t]
\hspace{-11px}
\includegraphics[width=1.04\linewidth]{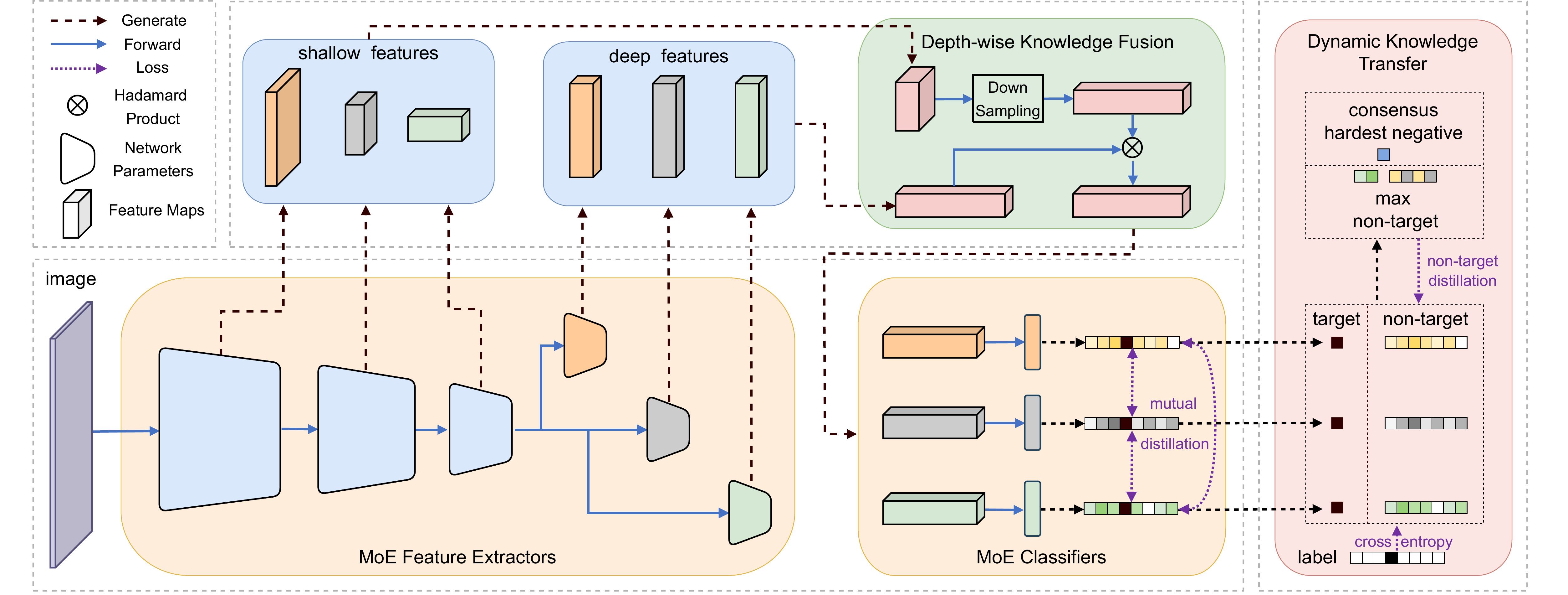}
\vspace{-10px}
\caption{The structure of the proposed SHIKE. Each expert in MoE fuses the features from its own exclusive layers (deep features) and ones from the shared layers (intermediate features). The fused features are then used for mutual and dynamic knowledge distillation for better model optimization.} 
\label{architecture}
\end{figure*}

\section{Methodology}
In this section, we introduce the proposed SHIKE in detail, which aims to enhance knowledge transfer in MoE. 
SHIKE contains two novel components named Depth-wise Knowledge Fusion (DKF) and Dynamic Knowledge Transfer (DKT). 
DKF aggregates knowledge diversely from experts with different model depths, and DKT is specifically designed for transferring knowledge among diverse experts.
The overall structure is shown in~\cref{architecture}.

\subsection{Preliminaries}
\label{Preliminaries}
We denote $\{x_i, y_i\}^N_{i=1}$ as all data points in the training set, where each sample $x_i$ has a corresponding label $y_i\in\{1,...,C\}$. 
The size of the training set is $N=\sum_{c=1}^C n_c$, where $n_c$ represents the number of training data in class $c$. Given a set of long-tailed data, the number of training data decreases according to the class indices, i.e., $n_1>n_2>...>n_C$. 
Long-tailed learning aims to build a deep model on such long-tailed data by treating each class equally important.

We suppose a deep neural network parameterized by $\theta$ contains $M$ experts. 
Usually, the network architecture of MoE makes the first several layers shared for all experts and the last few layers exclusive for each expert. 
Without loss of generality, we take ResNet~\cite{He2016resnet} as an example. 
We denote the shared layers of a ResNet model as $S$ stages for $M$ experts.
Only the last stage is adopted as the exclusive parameter for each expert. 
For expert $m$, we denote the parameters of its exclusive stage as $\theta^{m}_{S+1}$, which is then followed by a linear layer parameterized as $\varphi^{m}$. 
Given a data $x$, the intermediate features $\mathbf{f}_s$ from stage $s\left(1\leq s \leq S\right)$ of the shared network are calculated by
\setlength\abovedisplayskip{4pt}
\setlength\belowdisplayskip{4pt}
\begin{equation}
     \mathbf{f}_s = \theta_s \circ \cdot\cdot\cdot \circ \theta_2 \circ \theta_1(x).
\end{equation}
The operation $\circ$ indicates function composition: $h\circ {g}({x})={h}({g}({x}))$.
After the shared network, the output logits generated by expert $m$ are calculated by:
\begin{equation}
    \mathbf{z}^m = \varphi^{m}(\mathbf{f}_{S+1}^m),
\end{equation}
where $\mathbf{f}_{S+1}^m=\theta^m_{K+1}\left(\mathbf{f}_S\right)$ represents the exclusive features extracted by expert $m$, and $\mathbf{f}_S$ represents the features extracted by the last shared stage of the network.
In this MoE framework, we denote exclusive features $\mathbf{f}_{S+1}^m$ as high-level features and their preceding features $\mathbf{f}_{s}, s=1,...,S$, as intermediate features.
During training, the cross-entropy loss can be calculated for each expert after obtaining the softmax probabilities.
For model inference, the logits are summed up among all experts for each class, and the class with the maximum one is regarded as the MoE model prediction.
\subsection{Depth-Wise Knowledge Fusion}
Knowledge distillation is a commonly adopted optimization strategy for MoE methods in long-tailed learning \cite{wang2020long, xiang2020learning, li2021self, li2022nested}.
However, these methods mainly focus on distillation from logits rather than intermediate features.
In the field of knowledge distillation, most state-of-the-art methods mainly take a feature-based manner where intermediate feature plays an essential part \cite{romero2014fitnets, zagoruyko2016paying}. 
This is a lack of considering the intermediate features, especially in MoE-based methods where all experts often share the same part of the network. 
It has been shown in \cref{motivation} that features from different depths of the network can provide comparative performance towards different parts of long-tailed distribution: deep features exhibit promising performance on the head classes while shallow features can be more effective for some tail classes. 

Therefore, to fully utilize the intermediate features during knowledge distillation in an MoE framework, we propose Depth-wise Knowledge Fusion (DKF) to aggregate features from different depths of a shared network with the high-level features extracted from each expert. 
For simplification, we assume the number of experts $M$ is less than or equal to the number of shared stages, namely $M \leq S$ so that each expert can utilize the intermediate features from a unique depth in the network.
Then, we can assign $M$ sets of intermediate features among $\mathbf{f}_s$, $s=1,...,S$ to each expert.
As different intermediate features have different sizes, one expert cannot simply concatenate or multiply them with the assigned features  $\mathbf{f}_s$ directly.
Therefore, we add several convolution layers for downsampling according to the depth of the intermediate features, achieving feature alignment between $\mathbf{f}_s$ and high-level features $\mathbf{f}_{S+1}^m$ extracted by expert $m$. 
Suppose the intermediate features after alignment are $\hat{\mathbf{f}}_s$. 
In DKF, we propose to fuse the intermediate features with high-level features by multiplication and then transform them into logits by $\varphi^{m}$:
\begin{equation}
      \mathbf{z}^m = \varphi^{m}\left(\hat{\mathbf{f}}_s \otimes \mathbf{f}_{S+1}^m\right),
\end{equation}
where $\otimes$ is the Hadamard product.
As shown in \cref{architecture}, the intermediate features from a stage are assigned to an expert and aggregated with the high-level features of this expert. 

To fully use the diverse features in DKF, we can apply knowledge distillation between any two experts to make them learn from each other.
As each expert in MoE often has the same architecture located in the deepest position of the network, it can be guaranteed that each expert can play the role of either a teacher or a student.
This enables mutual knowledge distillation between any two experts and provides a perfect opportunity for experts to aggregate different depths of knowledge:
\setlength\abovedisplayskip{3pt}
\setlength\belowdisplayskip{3pt}
\begin{equation}
    \mathcal{L}_{mu} = \displaystyle \sum_{j=1}^{M}\displaystyle\sum_{k\neq j}^M KL(\mathbf{p}^j|\mathbf{p}^k),
    \label{mu_loss}
\end{equation}
where $\mathbf{p}^j$ and $\mathbf{p}^k$ represent the softmax probabilities of class $j$ and $k$, respectively. 
This guarantees the knowledge transferring comprehensively between any two experts. 

Feature fusion with mutual knowledge distillation in this way has two main advantages: 
(1) It dynamically fuses intermediate information from different depths of the network with semantic information from experts, which implicitly assigns different preferences of long-tailed distribution to experts without intuitive severance. 
(2) With more low-level information being aggregated, logit-based knowledge distillation can be more effective since each expert's output has more diversity corresponding to different depths of the model.
\subsection{Dynamic Knowledge Transfer}
The effectiveness of knowledge distillation largely relies on the non-target logits, i.e., the logits do not belong to the target class $y$, which provides similar semantic information in addition to the target logit.
It is especially useful for long-tailed learning because the target logits of samples in the tail classes are usually relatively small during training such that the non-target logits can provide a comparable amount of information with the target label.
With DKF, the non-target logits of different experts are more diverse because each expert can extract features with different semantic information due to different model depths.
However, one circumstance during knowledge distillation may happen under long-tailed distribution that needs to be taken into consideration carefully.
The model will be biased towards the head classes such that some samples in the tail classes will have high prediction confidence on the head classes, especially when they share similar semantic features.
The non-target classes with high confidence logits are called hard negative classes \cite{lin2017focal, li2022nested}.
It is dangerous to conduct knowledge distillation if the experts have a consensus on the hardest negative class, which may be a head class, for the tail class samples because the misleading information may be transferred.

Based on the analysis above, we propose Dynamic Knowledge Transfer (DKT) to address the issue of the hardest negative class during knowledge distillation with the proposed DKF.
In addition to using the logits of all classes by the cross-entropy loss, DKT considers only non-target predictions from all experts and dynamically elects a teacher among them to handle the hardest negative class.  
For a sample $x$ with label $y$, its corresponding output logits of expert $m$ is $\mathbf{z}^m=[z^m_1,z^m_2,...,z^m_C]$. 
Following \cite{zhao2022decoupled}, we first decouple the logits into a target logit $z^m_y$ and non-target logits $[z_{I_1}^{m},z_{I_2}^m,...,z_{I_{C-1}}^m]$, where $\mathcal{I}=[I_1, I_2, ..., I_{C-1}]$ stores the index of non-target classes.
After logits decoupling, we introduce the non-target set to a new knowledge distillation problem with $C-1$ classes. 
The average logits of all experts can be calculated for each non-target class $[\bar{z}_{I_1},\bar{z}_{I_2},...,\bar{z}_{I_{C-1}}]$, where
\begin{equation}
    \Bar{z}_{I_i} = \frac{1}{M}\sum_{m=1}^{M}z^m_{I_i},
\label{nt_avg}
\end{equation}
for $i=1,...,C-1$.
We can thus identify $\max_i\{\Bar{z}_{I_i}\}$ among all non-target classes as the consensus hardest negative class, which is believed as the hardest negative class by joint prediction of MoE.
To effectively suppress the logit of consensus hardest negative class through softmax suppression, a teacher who can comprehensively utilize the non-target knowledge is needed. 
Specifically, DKT chooses the maximum non-target logit among all experts denoted as $\hat{z}_{I_i}$: 
\begin{equation}
\hat{z}_{I_i} = \max_{m=1,...,M}\{z^m_{I_i}\},
\end{equation}
for $i=1,...,C-1$.
The large values of the maximum non-target logits can effectively suppress the value of the consensus hardest negative logit after softmax on the $C-1$ non-target classes.
Taking advantage of the diversity in DKF, high values may appear on different non-target logits with different experts.
Therefore, the maximum non-target logits can dynamically suppress the consensus hardest negative logit after softmax among $C-1$ non-target classes.
Combining the consensus hardest negative with the maximum non-target logits, we can form a set of non-target logits called grand teacher:
\begin{equation}
    z^\mathcal{T}_{I_i}=\left\{ 
    \begin{array}{ll}
        \bar{z}_{I_i}, & i=\argmax_{j}\{\bar{z}_{I_j}\},\\
        \hat{z}_{I_i}, & \mathrm{otherwise},\\
    \end{array}
\right.
\label{avg_nt2}
\end{equation}
for $i=1,...,C-1$. 
Note that the grand teacher is only for suppressing the hardest negative class within non-target classes while the target class is not involved.
After electing the grand teacher, non-target knowledge distillation is performed between it and each expert. 
The non-target probabilities for grand teacher and students are calculated by the following formulation:
\vspace{+2px}
\begin{equation}
    \widetilde{p}^\mathcal{T}_{I_i}=\frac{\exp (z_{I_i}^\mathcal{T})}{\sum_{i=1}^{C-1} \exp (z_{I_i}^\mathcal{T})},\ \ \ \ \widetilde{p}^m_{I_i}=\frac{\exp (z_{I_i}^m)}{\sum_{i=1}^{C-1} \exp (z_{I_i}^m)},
\label{equa:nt_probabilities}
\vspace{+2px}
\end{equation}
for $i=1,...,C-1$ and $m=1,...,M$. Therefore, the knowledge distillation for non-target logits among SHIKE's experts can be formulated as:
\begin{equation}
    \mathcal{L}_{nt} = \displaystyle \sum_{m=1}^{M} KL(\widetilde{\mathbf{p}}^\mathcal{T}|\widetilde{\mathbf{p}}^m).
    \label{nt_loss}
\end{equation}
For a particular sample, the hardest negative of its corresponding outputs may vary not only among experts but also along the training process. 
DKT can dynamically choose the hardest negative among experts and reduce its probability without affecting the target logit.
\subsection{Overall Training Paradigm}
SHIKE adopts a decoupled training scheme that optimizes the feature extractor and classifier separately, as it has been shown that class-balanced joint training strategies for long-tailed data may hurt representation learning \cite{kang2019decoupling, zhou2020bbn}.
For feature extractor training, we adopt both mutual knowledge distillation loss $\mathcal{L}_{mu}$ in Eq. (\ref{mu_loss}) by DKF and non-target knowledge distillation loss $\mathcal{L}_{nt}$ in Eq. (\ref{nt_loss}) by DKT.
Meanwhile, to preserve the information from the original distribution untouched for representation learning, vanilla cross-entropy loss $\mathcal{L}_{ce}$ is also applied for each expert.
Therefore, the above three loss functions during the representation learning stage are assembled as a whole optimization objective:
\begin{equation}
    \mathcal{L} = \mathcal{L}_{ce} + \alpha\mathcal{L}_{nt} + \beta\mathcal{L}_{mu},
    \label{equa:feat_learning}
\end{equation}
where $\alpha$ and $\beta$ are trade-off hyperparameters.
For classifier training, the goal is to train a balanced classifier with the feature extractor frozen. 
We utilize the balanced softmax cross entropy (BSCE)~\cite{ren2020balanced} as the loss function $L_{bsce}$ to simply optimize a new classifier for each expert:
\begin{equation}
    \mathcal{L}_{bsce} = \displaystyle - \sum_{m=1}^M \log\left(\frac{n_{y} \exp \left(z^m\right)}{\sum_{j=1}^C n_{j} \exp \left(z_j^m\right)}\right).
\label{bsce_loss}
\end{equation}  
Knowledge distillation is not considered in the stage of classifier re-training as it will encourage classifiers to be similar, which is harmful to experts to make joint predictions.

\section{Experiments}
\subsection{Datasets}
{\bf CIFAR-100-LT} is a subset of the original CIFAR-100~\cite{krizhevsky2009learning} with long-tailed distribution. 
The imbalance factor of CIFAR-100-LT can be set at 50 or 100, where 100 means that the largest class contains 100 times of samples than the smallest class. 
The validation set remains the same as the original CIFAR-100 with 10,000 samples in total. 
{\bf ImageNet-LT} is also a subset of the original ImageNet~\cite{deng2009imagenet}, which is first proposed in~\cite{liu2019large}. 
It has an imbalance factor of 256 following Pareto distribution with a power value of 0.6. 
With 1,000 classes in total, the training set and test set contain 115.8K samples and 50K samples, respectively.
{\bf iNaturalist 2018} \cite{van2018inaturalist} is a large-scale real-world dataset. 
It is extremely imbalanced with 437.5K samples from 8,142 categories.
{\bf Places-LT} is created from the large-scale dataset Places~\cite{zhou2017places} with 184.5K samples from 365 categories.

\subsection{Implementation Details}
For CIFAR100-LT, we use ResNet-32 as the backbone. AutoAugment~\cite{cubuk2019autoaugment} and Cutout~\cite{devries2017improved} are adopted by following~\cite{cui2021parametric, ren2020balanced}. 
For ImageNet-LT, ResNet-50 and ResNeXt-50 (32x4d) are adopted. 
Similarly, we use ResNet-50 and ResNet-152 for iNaturalist 2018 and Places-LT. 
The above four datasets are trained with learning rates of 0.05, 0.2, 0.025, and 0.02, respectively. 
All models are trained for 180 epochs except Places-LT with 30 epochs of fine-tuning as it utilizes a pre-trained model. 
If not specified, we adopt SGD optimizer with momentum 0.9, cosine schedule of decaying to 0, and weight decay of 5e-4 for all experiments.
RandAugment~\cite{cubuk2020randaugment} is used for ImageNet-LT and iNaturalist-2018 by following~\cite{li2022nested}.
Also, during the classifier training phase, the cosine learning rate scheduler restarts, and classifiers of experts are trained for 20 more epochs.

\begin{table}[tp]\small
\centering
\setlength{\tabcolsep}{8pt}
\renewcommand{\arraystretch}{1.}
\begin{tabular}{l|c|c|c}
\hline
\multicolumn{1}{l|}{\multirow{2}{*}{Method}} & \multicolumn{1}{l|}{\multirow{2}{*}{Year}} & \multicolumn{2}{c}{CIFAR100-LT}                                \\ \cline{3-4} 
        &   & 100 & 50  \\ \hline
\multicolumn{4}{l}{\emph{Single model}}  \\ \hline
Focal Loss~\cite{lin2017focal}  & 2017  & 42.3  & - \\
OLTR~\cite{liu2019large} & 2019 & 43.4  & -    \\
LDAM-DRW~\cite{Cao2019Learning} & 2019   & 44.4  & -   \\
$\tau$-norm~\cite{kang2019decoupling}   & 2020 & 45.4  & -    \\
cRT~\cite{kang2019decoupling} & 2020  & 45.6 & -    \\
BALMS~\cite{ren2020balanced} & 2020 & 50.7   & -    \\
LADE~\cite{hong2021disentangling}  & 2021 &  45.4  & 50.5  \\
GCL~\cite{li2022long} & 2022  & 48.7  & 53.6 \\
Weight Balancing~\cite{alshammari2022long} & 2022 & 53.6 & 57.7\\ \hline
\multicolumn{4}{l}{\emph{Contrastive \& Hybrid methods}}  \\ \hline
BALMS+BatchFormer~\cite{hou2022batchformer}  & 2022 & 51.7  & -   \\
PaCo~\cite{cui2021parametric}  & 2021 & 51.9 & 56.0 \\
PaCo+BatchFormer~\cite{hou2022batchformer} & 2022 & 52.4 & -  \\
BCL~\cite{zhu2022balanced} & 2022 & 52.0 & 56.6   \\ \hline
\multicolumn{4}{l}{\emph{MoE-based methods}}\\ \hline
RIDE (3E)~\cite{wang2020long}  & 2021 & 48.3  & -  \\
ResLT (3E)~\cite{cui2022reslt} & 2022 & 45.3  & 50.0  \\
TLC (4E)~\cite{li2022trustworthy}  & 2022 & 50.1  & -  \\
NCL (S)~\cite{li2022nested} & 2022 &53.3  & 56.8\\
NCL (3N)~\cite{li2022nested} & 2022 & 54.2 & 58.2  \\ \hline
Ours (3E) & - & \textbf{56.3} & \textbf{59.8} \\ \hline
\end{tabular}
\vspace{-5px}
\caption{Comparison results on CIFAR100-LT with imbalance factor of 100 and 50.  }
\label{comparison_cifar}
\vspace{-10px}
\end{table}

\begin{table}[t]\small
\centering
\resizebox{0.95\linewidth}{!}{
\begin{tabular}{l|c|ccc}
\hline
\multirow{2}{*}{Method} & \multirow{2}{*}{Year} & \multicolumn{2}{c|}{ImageNet-LT}  &  iNat  \\ 
\cline{3-5}  &   &\multicolumn{1}{c|}{R-50} & \multicolumn{1}{c|}{RX-50} & R-50  \\ \hline
\multicolumn{5}{l}{\emph{Single model}}  \\ \hline
OLTR~\cite{liu2019large} & 2019 & -  & -  & 63.9  \\
LDAM-DRW~\cite{Cao2019Learning} & 2019   & - & - & 68.0  \\
cRT~\cite{kang2019decoupling} & 2020  & 47.3 & 49.6  &65.2 \\
$\tau$-norm~\cite{kang2019decoupling}   & 2020 & 46.7  & 49.4 & 65.6  \\
BALMS~\cite{ren2020balanced}  & 2020 &50.1 & -  & -  \\
LA~\cite{menon2020long} & 2021 & -  & - &66.4    \\   %
CAM~\cite{Zhang2021Bag} & 2021 & -  & - &70.9    \\
GCL~\cite{li2022long} & 2022  & 54.9  & - &72.0 \\
Weight Balancing~\cite{alshammari2022long} & 2022 & - & 53.9 & 70.2\\ \hline
\multicolumn{4}{l}{\emph{Contrastive \& Hybrid methods}}  \\ \hline
SSD~\cite{li2021self} & 2021 & -  & 56.0  &- \\
PaCo~\cite{cui2021parametric} & 2021 &  57.0  & 58.2  & 73.2 \\
BCL~\cite{zhu2022balanced} & 2022  & 56.0 & -  & 71.8 \\ 
RIDE+BF~\cite{hou2022batchformer} & 2022 & 55.7 & -  & 74.1 \\   
BALMS+BF~\cite{hou2022batchformer}  & 2022 & 51.1 & -  & - \\ \hline
\multicolumn{4}{l}{\emph{MoE-based methods}}\\ \hline
BBN~\cite{zhou2020bbn} & 2020 & 48.3  & 49.3  & 66.3 \\
RIDE (3E)~\cite{wang2020long} & 2021 & 55.4 & 56.8  & 72.6 \\ 
ACE~\cite{cai2021ace} & 2021 &54.7 & 56.6  & -  \\
NCL (S)~\cite{li2022nested} & 2022 & 57.4 & 58.4 & 74.2\\
NCL (3N)~\cite{li2022nested} & 2022 & 59.5 & \textbf{60.5}  & 74.9  \\ \hline
Ours (3E) & - & \textbf{59.7} & 59.6 & \textbf{75.4} \\ \hline

\end{tabular}
}
\vspace{-5px}
\caption{Comparison results on ImageNet-LT and iNaturalist 2018 (iNat). R-50 and RX-50 are short for ResNet-50 and ResNeXt-50 (32x4d), respectively.}
\label{comparison_img_inat}
\vspace{-10px}
\end{table}

\begin{figure*}[t!]
  \centering
  \begin{subfigure}{0.646\linewidth}
     \includegraphics[width=1.0\linewidth]{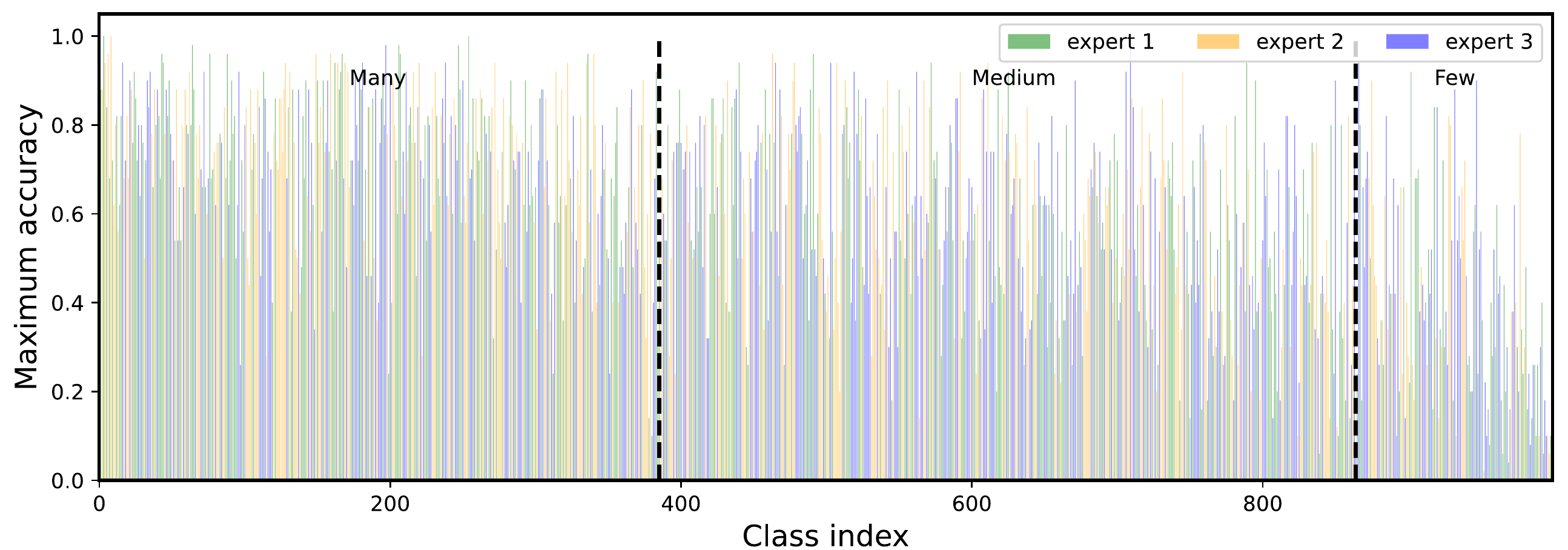}
    \caption{Maximum class accuracy among experts.}
    \label{fig:short-a}
  \end{subfigure}
  \hfill
  \begin{subfigure}{0.314\linewidth}
         \includegraphics[width=1.0\linewidth]{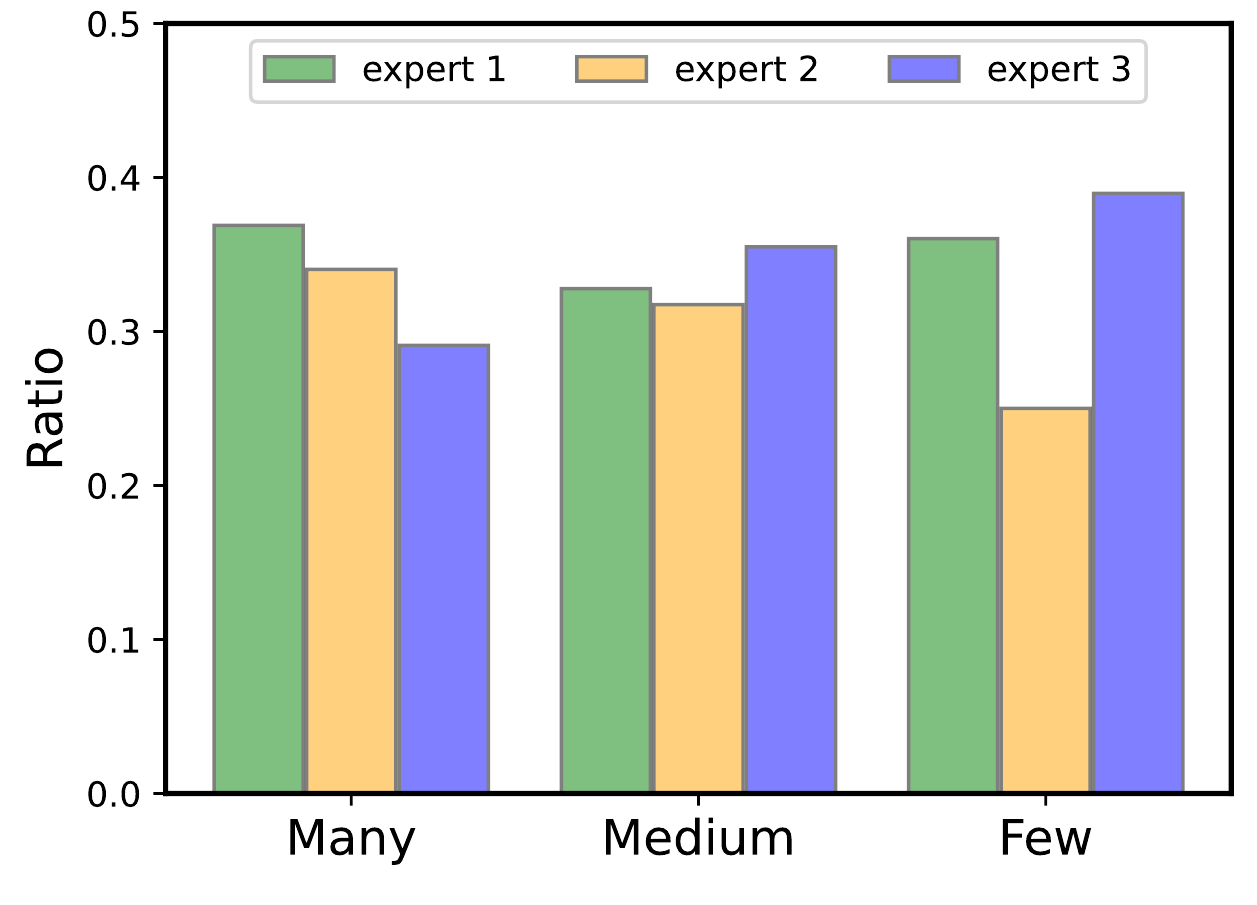}
    \caption{Ratio of classes with the highest accuracy.}
    \label{fig:short-b}
  \end{subfigure}
  \vspace{-8px}
  \caption{Preferences of different experts in SHIKE.
    (a) The highest accuracy among experts is shown for each class on the test set of ImageNet-LT.
    (b) We calculate the ratio of class numbers that each expert is most skilled at within three divisions.
    The experiment is conducted with ResNet-50 and the number of experts is set to 3.}
  \label{preference_histogram}
  \vspace{-15pt}
\end{figure*}

\subsection{Main Results}
All comparison results with other state-of-the-art methods for long-tailed learning are presented in \cref{comparison_cifar}-\ref{comparison_places}. 
We compare the proposed SHIKE with single model methods, contrastive methods, hybrid methods and MoE-based methods.
All of them are proposed for long-tailed data. 
Specific settings for MoE-based models are marked in parentheses: S for the single model, N for the full network, and E for the expert. The best results are presented in \textbf{bold}.
\vspace{-15px}
\paragraph{CIFAR100-LT} 
The comparison results on CIFAR100-LT with the imbalance factor of 100 and 50 are shown in \cref{comparison_cifar}. 
Note that the adopted ResNet-32 model has only three stages, which provide two shared stages to generate intermediate features for fusion and one exclusive stage for each expert.
In this case, we let two experts fuse with the same shallowest intermediate features. 
We group the existing methods based on their types.
SHIKE achieves better performance within or out of MoE-based methods. 
For example, it achieves 2.1\% and 1.6\% improvements over the second-best method NCL (3N) on the imbalance factor of 100 and 50, respectively.
Besides, it is worth noting that in the previous state-of-the-art methods, NCL adopts three complete networks as its experts, while SHIKE only utilizes experts consisting of the last stage of the ResNet along with one or two downsampling layers.
Another advantage of the proposed SHIKE is we only train it for 200 epochs, which is less than the method following a contrastive learning strategy that requires more epochs.
\vspace{-15pt}
\paragraph{ImageNet-LT and iNaturalist 2018.} 
We report overall Top-1 accuracy for ImageNet-LT and iNaturalist 2018 in~\cref{comparison_img_inat}. 
For a fair comparison, we use SHIKE to train MoEs 200 epochs for both datasets.
For ImageNet-LT, SHIKE can perform better than all the competing methods, including NCL~\cite{li2022nested}. 
It is worth noting that NCL trains its MoE for 400 epochs with contrastive training strategies.
Moreover, NCL adopts three whole networks as experts, consuming more computational overhead for training. 
Our method is less computationally expensive but still achieves comparable or even better performance than NCL. 
We also conduct an experiment with 400 epochs for ResNet-50 on ImageNet-LT, achieving a performance of 60.3\%.
\begin{table}[tb]\small
\centering
\resizebox{\linewidth}{!}{
\begin{tabular}{l|l|cccc}
\hline
\multirow{2}{*}{Method} & \multirow{2}{*}{Year} & \multicolumn{4}{c}{Places-LT} \\ \cline{3-6} 
                        &  & Many   & Med  & Few   & All  \\ \hline
\multicolumn{6}{l}{\emph{Single model}}  \\ \hline
Focal Loss~\cite{lin2017focal}  & 2017 & 41.1 & 34.8 & 22.4 & 34.6  \\
OLTR~\cite{liu2019large} & 2019 & \textbf{44.7} & 37.0 & 25.3 & 35.9  \\
NCM~\cite{kang2019decoupling}  & 2020 & 40.4 & 37.1 & 27.3 & 36.4  \\
cRT~\cite{kang2019decoupling}  & 2020 & 42.0 & 37.6 & 24.9 & 36.7  \\
$\tau$-norm ~\cite{kang2019decoupling} & 2020 & 37.8 & 40.7 & 31.8 & 37.9  \\
LWS~\cite{kang2019decoupling} & 2020 & 40.6 & 39.1 & 28.6 & 37.6  \\
BALMS~\cite{ren2020balanced} & 2020 & 41.2 & 39.8 & 31.6 & 38.7  \\
LADE~\cite{hong2021disentangling} & 2021 & 42.8 & 39.0 & 31.2 & 38.8  \\
DisAlign~\cite{zhang2021distribution}  & 2021 & 40.4 & 42.4 & 30.1 & 39.3  \\
GCL~\cite{li2022long}   & 2022 & - & - & - & 40.6  \\
\hline
\multicolumn{6}{l}{\emph{Contrastive \& Hybrid methods}}  \\ \hline
LDAM+RSG~\cite{wang2021rsg} & 2021 & 41.9 & 41.4 & 32.0 & 39.3  \\
PaCo~\cite{cui2021parametric} & 2021 & 37.5 & \textbf{47.2} & 33.9 & 41.2  \\\hline
\multicolumn{6}{l}{\emph{MoE-based methods}}\\ \hline
LFME~\cite{xiang2020learning} & 2020 & 39.3 & 39.6 & 24.2 & 36.2  \\
NCL (S)~\cite{li2022nested} & 2022 & - & - & - & 41.5   \\
NCL (3N)~\cite{li2022nested} & 2022 &- & - & - & 41.8 \\ \hline
Ours (3E) & - & 43.6 & 39.2 & \textbf{44.8} & \textbf{41.9} \\ \hline
\end{tabular}}
\vspace{-5px}
\caption{Comparison results on Places-LT. The results are shown by different class divisions (Many, Medium, and Few) as well as the overall accuracy (All).}
\label{comparison_places}
\vspace{-10px}
\end{table}
\vspace{-15pt}
\paragraph{Places-LT.} 
As previous works consider the pre-trained backbone of ResNet-152 as a whole, it is troublesome for SHIKE to implement the model. 
To demonstrate the effectiveness of the proposed SHIKE, we keep the shared part of the model fixed after loading the pre-trained parameters. 
Then, we only fine-tune the exclusive part within experts and its corresponding downsampling layers, which are crucial for the proposed component DKF. 
The comparison results are listed in~\cref{comparison_places}. 
This experiment shows that intermediate features from the pre-trained model can also help to boost the overall performance on long-tailed visual recognition.
To further reveal the effectiveness of SHIKE, we report the accuracy on three divisions of the classes, namely many-shot classes ($>$100 training samples), medium-shot classes (20$\sim$100 training samples) and few-shot classes ($<$20 training samples).
In addition to achieving slightly higher performance than the state-of-the-art, we find it more fascinating that SHIKE achieves 44.8\% accuracy on the few-shot classes, which is more than 10\% higher than the runner-up 33.9\% achieved by PaCo~\cite{cui2021parametric}.
As a result, SHIKE has a more balanced test performance compared to the contrastive learning methods.
\subsection{Ablations and Model Validation}
\begin{figure}[!t]
    \vspace{-12px}
    \centering
    \hspace{-8pt}
    \includegraphics[width=0.9\linewidth]{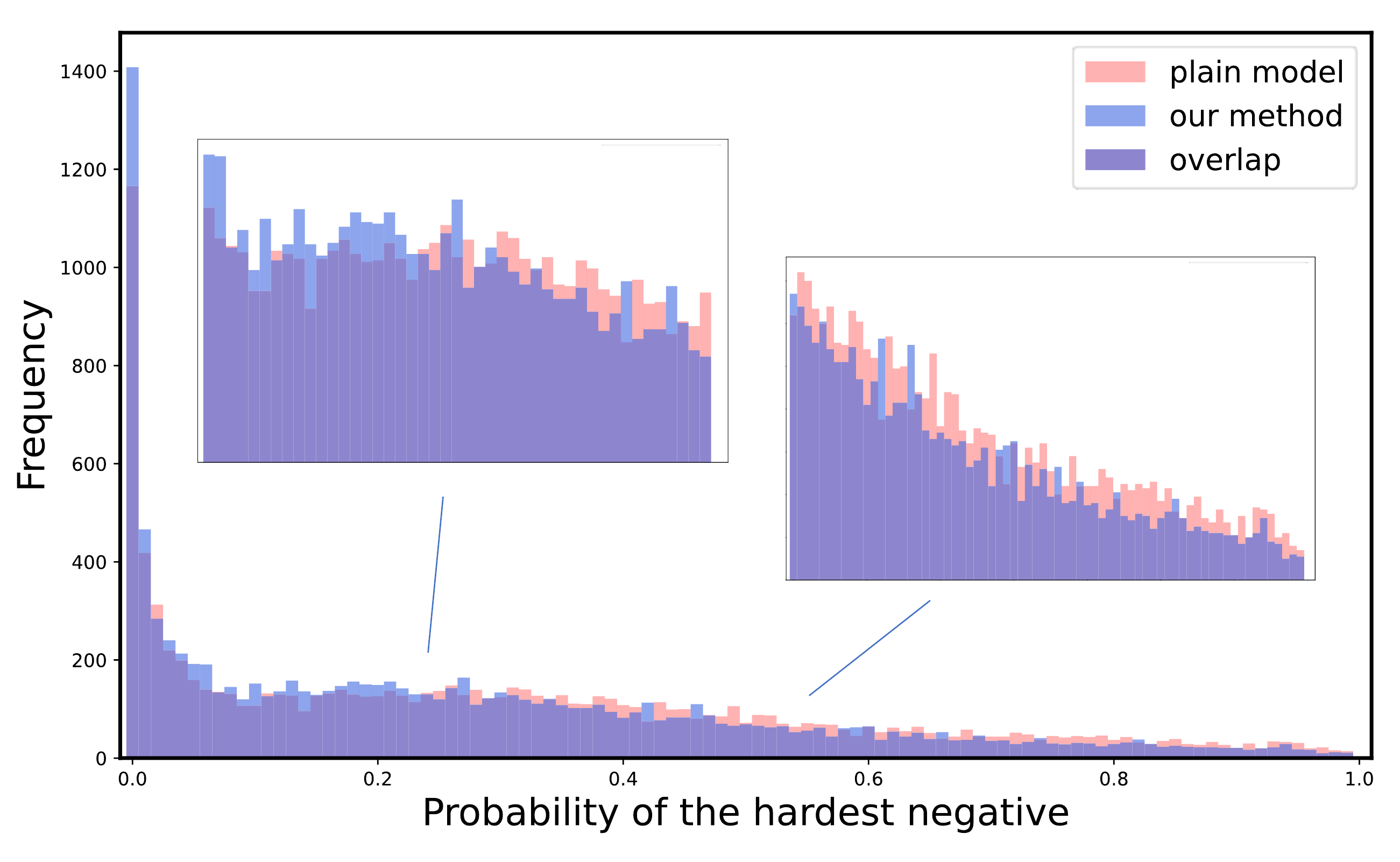}
    \vspace{-8px}
    \caption{Probability distribution of the hardest negative class for models trained on CIFAR100-LT with an imbalance factor of 100. 
    The result is counted on the test set of CIFAR100-LT to show the effectiveness of suppressing the hardest negative class by DKT.
    }
    \label{hardest_negative}
\vspace{-10px}
\end{figure}

\begin{table}[tb]\small
\centering
\setlength{\tabcolsep}{8pt}
\renewcommand{\arraystretch}{1.}
\begin{tabular}{llll|l}
\hline
MoE        & DKF        & $\mathcal{L}_{mu}$ & $\mathcal{L}_{nt}$ & Acc            \\ \hline
           &            &                     &                    & 50.04          \\
\checkmark &            &                     &                    & 54.23          \\
\checkmark & \checkmark &                     &                    & 55.26          \\
\checkmark & \checkmark & \checkmark          &                    & 55.59          \\
\checkmark & \checkmark &                     & \checkmark         & 55.79          \\
\checkmark &            & \checkmark          & \checkmark         & 54.82          \\
\checkmark & \checkmark & \checkmark          & \checkmark         & 
\textbf{56.34} \\ \hline
\end{tabular}
\vspace{-5px}
\caption{Ablation study on the effects of different components in the proposed SHIKE. The experiment is conducted on CIFAR100-LT with an imbalance factor of 100.}
\label{ablation_study}
\vspace{-10pt}
\end{table}
\paragraph{Ablation Studies on Components of SHIKE.} 
Shown in~\cref{ablation_study}, the ablation study is conducted on CIFAR100-LT with an imbalance factor of 100. 
Key components of SHIKE are further subdivided into MoE, DKF, and two loss functions $\mathcal{L}_{mu}$ and $\mathcal{L}_{nt}$. 
MoE means whether the method uses three experts or a single plain ResNet-32.
All experiments are conducted following a decoupled training scheme, and accuracy is calculated on the balanced test set. 
The accuracy for a single plain model without any component is 50.04\%, which acts as a baseline for ablation.
When applying the MoE architecture, the accuracy is significantly boosted to 54.23\%, which validates the effectiveness of experts' ensemble toward long-tailed learning.  
When applying DKF to MoE, the performance can be boosted by around 1\%.
Based on DKF, the mutual knowledge distillation loss $\mathcal{L}_{mu}$ and the non-target knowledge distillation $\mathcal{L}_{mt}$ (DKT) can further improve the accuracy by 0.33\% and 0.53\%, respectively.
When all the components are applied, we achieve the highest accuracy of 56.34\%, which is around 2\% higher than the plain MoE model.
An interesting observation is that if we do not apply DKF, all the other components can only bring 0.59\% improvement.
This proves that the DKF is the fundamental architecture in SHIKE, which assigns more meaningful features to experts in MoE for further knowledge distillation and model optimization.
\vspace{-15pt}
\paragraph{Evaluation on the Hardest Negatives.}
The ablation study in \cref{ablation_study} has shown the effectiveness of DKT in terms of accuracy. 
To further validate how DKT suppresses the hardest negatives, we conduct an evaluation based on the test set. 
A single model and the proposed SHIKE are utilized, and both model's classifiers are trained with BSCE along with keeping the feature extractor fixed. 
As shown in~\cref{hardest_negative}, we can see that the number of the hardest negatives with large values is reduced, which indicates that SHIKE can effectively alleviate the influence of hardest negatives during knowledge distillation in MoE.

\vspace{-15pt}
\paragraph{Evaluation on the Preference of Experts.}
To show the preferences of experts for long-tailed distribution, an experiment is conducted on the test set of ImageNet-LT, and the experts are evenly assigned with three different levels of intermediate features from shallow to deep accordingly.
\cref{preference_histogram} (a) shows the highest accuracy among experts for each class.
It can be observed that the diversity among the three experts is quite high, where each expert performs well in different classes that are distributed from head to tail of the distribution.
We also calculate the ratio of classes with the highest accuracy among experts.
From~\cref{preference_histogram}(b), we can see that expert 1 and 3 with the shallowest and the deepest intermediate features performs better on few-shot division.
While expert 2, which is assigned by the middle intermediate features, follows a normal accuracy distribution which is consistent with long-tailed distribution.
Moreover, for classes in the many-shot division, the shallowest feature is superior in helping expert to achieve higher performance.
In all, experts with different intermediate features appear to have different preferences for long-tailed distribution, which makes experts skilled at different parts of the distribution.


\begin{table}[tb]\small
\vspace{-8px}
\centering
\setlength{\tabcolsep}{8pt}
\renewcommand{\arraystretch}{1.}
\begin{tabular}{c|ccc}
\hline
$E$     & \multicolumn{3}{c}{Depth arrangement} \\ \hline
\multirow{2}{*}{1} & A         & B        & C        \\              
                       & 54.10        & 54.75       & \textbf{55.84}         \\ \hline
\multirow{2}{*}{2} & A B         & B C        & A C        \\              
                   & 57.39        & \textbf{58.79}       & 58.62         \\ \hline
\multirow{2}{*}{3} & \multicolumn{3}{c}{A B C}             \\
                   & \multicolumn{3}{c}{\textbf{59.72}}              \\ \hline
\multirow{2}{*}{4} & {\footnotesize A B C A}   &  {\footnotesize A B C B}    &  {\footnotesize A B C C}    \\
                   & 59.83        & \textbf{59.90}          &   59.77         \\ \hline
\end{tabular}
\vspace{-5px}
\caption{Ablation study on the effects of expert number in the proposed SHIKE. The experiment is conducted on ImageNet-LT.}
\label{experts_num}
\vspace{-10pt}
\end{table}

\vspace{-15pt}
\paragraph{Evaluation on the Number of Experts.}
To show the influence of the number of experts, we conduct an experiment on ImageNet-LT by varying the number of experts from 1 to 4 and different depth combinations. 
As shown in~\cref{experts_num}, letters from A to C represent depths from shallow to deep.  
The overall accuracy of the ensemble generally rises along with the increasing number of experts.
Moreover, it shows that as the number of experts grows, architectures with more heterogeneous experts promote more.

\section{Conclusion}
We have proposed a Self-Heterogeneous Integration with Knowledge Excavation (SHIKE) for long-tailed visual recognition. 
The proposed SHIKE consists of Depth-wise Knowledge Fusion (DKF) and Dynamic Knowledge Transfer (DKT). 
DKF fuses the depth-wise intermediate features with high-level features and thereby provides more informative features for experts to accommodate the long-tailed distribution. 
DKT exploits the non-target knowledge among diversified experts to reduce the hardest negative for representation learning, which can further improve the performance on the tail classes.
Extensive experiments have been conducted on four benchmarks and SHIKE achieved excellent performance compared to state-of-the-art counterparts.

\vspace{-10pt}
\paragraph{Acknowledgements}
This work was supported by the National Natural Science Foundation of China under Grants 62002302 and U21A20514, the FuXiaQuan National Independent Innovation Demonstration Zone Collaborative Innovation Platform under Grant 3502ZCQXT2022008, NSFC/Research Grants Council (RGC) Joint Research Scheme under Grant N\_HKBU214/21, the General Research Fund of RGC under Grants 12201321 and 12202622, the Natural Science Foundation of Fujian Province under Grant 2020J01005.

\clearpage
{\small
\bibliographystyle{ieee_fullname}
\bibliography{egbib}
}

\end{document}